\documentclass[]{interact}

\usepackage{epstopdf}% To incorporate .eps illustrations using PDFLaTeX, etc.
\usepackage[caption=false]{subfig}% Support for small, `sub' figures and tables

\usepackage{graphicx}      % include this line if your document contains figures
\usepackage{multirow}
\usepackage{pbox}

\usepackage[T1]{fontenc}
\usepackage{lmodern}

\usepackage{url}

\usepackage[acronym]{glossaries}
\newacronym{das}{DAS}{Driver Assistance Systems}
\newacronym{rl}{RL}{Reinforcement Learning}
\newacronym{vdc}{VDC}{Vehicle Dynamic Controls}
\newacronym{nn}{NN}{Neural Network}
\newacronym{drl}{DRL}{Deep Reinforcement Learning}
\newacronym{nmpc}{NMPC}{Nonlinear Model Predictive Controller}
\newacronym{nlp}{NLP}{Nonlinear Optimization Problem}
\newacronym{mpc}{MPC}{Model Predictive Controller}
\newacronym{cnn}{CNN}{Convolutional Neural Network}
\newacronym{cte}{CTE}{Cross Track Error}
\newacronym{ddpg}{DDPG}{Deep Deterministic Policy Gradient}
\newacronym{swa}{SWA}{Steering Wheel Angle}
\newacronym{tbp}{TBP}{Throttle/Brake Position}
\newacronym{a4ws}{A4WS}{Active Four Wheel Steering}
\newacronym{a4wd}{A4WD}{Active Four Wheel Drive}
\newacronym{as}{AS}{Active Suspension}
\newacronym{ppo}{PPO}{Proximal Policy Optimization}
\newacronym{a3c}{A3C}{Asynchronous Advantageous Actor Critic}
\newacronym{ai}{AI}{Artificial Intelligence}
\newacronym{ml}{ML}{Machine Learning}
\newacronym{abs}{ABS}{Anti Block System}
\newacronym{asr}{ASR}{Anti Slip Regulation}
\newacronym{mlp}{MLP}{Multilayer Perceptron}
\newacronym{dil}{DiL}{Driver in the Loop}
\newacronym{os}{OS}{oversteer}
\newacronym{us}{US}{understeer}
\newacronym{4wd}{4WD}{Four Wheel Drive}

\usepackage{array}

\usepackage{makecell}

\usepackage[numbers,sort&compress]{natbib}% Citation support using natbib.sty
\bibpunct[, ]{[}{]}{,}{n}{,}{,}% Citation support using natbib.sty
\renewcommand\bibfont{\fontsize{10}{12}\selectfont}% Bibliography support using natbib.sty
\makeatletter% @ becomes a letter
\def\NAT@def@citea{\def\@citea{\NAT@separator}}% Suppress spaces between citations using natbib.sty
\makeatother% @ becomes a symbol again

\theoremstyle{plain}% Theorem-like structures provided by amsthm.sty

\theoremstyle{definition}

\theoremstyle{remark}

\begin{document}

% \articletype{ARTICLE TEMPLATE}% Specify the article type or omit as appropriate

\title{Self driving algorithm for an active four wheel drive racecar}

\author{
\name{Gergely Bári\textsuperscript{a} \thanks{CONTACT Gergely Bári Email: bairg82@gmail.com} and László Palkovics\textsuperscript{b}}
\affil{\textsuperscript{a}HUMDA Lab nonprofit Kft., 1113 Budapest, Dávid Ferenc utca 4-6. Hungary; \textsuperscript{b}Széchenyi István University, 9026 Győr, Hungary}
}

% \author{
% \name{A.~N. Author\textsuperscript{a}\thanks{CONTACT A.~N. Author. Email: latex.helpdesk@tandf.co.uk} and John Smith\textsuperscript{b}}
% \affil{\textsuperscript{a}Taylor \& Francis, 4 Park Square, Milton Park, Abingdon, UK; \textsuperscript{b}Institut f\"{u}r Informatik, Albert-Ludwigs-Universit\"{a}t, Freiburg, Germany}
% }

\maketitle

\begin{abstract}
%%shorter
% This study employs \acrfull{rl}, specifically \acrfull{ppo}, to train an agent for autonomous racecar driving, focusing on vehicle control at the grip limit. Critically, the agent directly outputs steering commands and independent torques for all four wheels, bypassing conventional pedal inputs and inherently solving the \gls{a4wd} control problem within the driving policy. Simulations demonstrate the RL agent achieves performance comparable to, and in some cases surpassing, traditional physics-based \gls{a4wd} controllers, exhibiting sophisticated optimization of wheel torques for enhanced handling and stability in grip-limited racing scenarios. This research underscores the potential of \acrshort{rl} to develop unified, high-performance control strategies, suggesting RL-based systems as a potent alternative for advancing autonomous driving and vehicle dynamics control.

Controlling autonomous vehicles at their handling limits presents significant challenges, particularly for electric vehicles equipped with active four-wheel drive (A4WD) systems offering independent wheel torque control. While traditional Vehicle Dynamics Control (VDC) methods rely on complex physics-based models and intricate coordination strategies, this study explores the potential of Deep Reinforcement Learning (DRL) to develop unified, high-performance controllers for such systems. We employ the Proximal Policy Optimization (PPO) algorithm to train an autonomous agent for optimal lap times in a simulated racecar environment (TORCS), focusing specifically on operation at the tire grip limit. Critically, the agent learns an end-to-end policy that directly maps real-vehicle-friendly states—eg.: velocities, accelerations, and yaw rate—to control outputs consisting of a steering angle command and, distinctively, independent torque commands for each of the four wheels. This formulation bypasses conventional pedal inputs and explicit torque vectoring algorithms, allowing the agent to implicitly learn the complex A4WD control logic required for maximizing performance and stability. Simulation results demonstrate that the RL agent successfully learns sophisticated control strategies, dynamically optimizing wheel torque distribution corner-by-corner to enhance handling characteristics and effectively mitigate the base vehicle's inherent understeer. The learned behaviors mimic and, in certain aspects of grip utilization, potentially surpass the capabilities of traditional physics-based A4WD controllers, achieving competitive lap times. This research underscores the significant potential of DRL to create adaptive, high-performance control systems for complex vehicle dynamics, suggesting RL-based approaches as a potent alternative for advancing autonomous driving capabilities, particularly in demanding, grip-limited scenarios relevant to both racing and road vehicle safety.

\end{abstract}


%------------------------------
\section{Introduction}
% Original start kept, slightly expanded motivation
The pursuit of fully autonomous driving systems continues to accelerate, driven by the promise of transformative improvements in traffic efficiency, road safety, and transportation accessibility. While significant progress has been made in structured environments like highways, a critical frontier remains: ensuring reliable vehicle control under highly dynamic conditions, particularly scenarios demanding operation near or at the limits of tire adhesion \cite{brown_coordinating_2020}. Mastering vehicle behavior in these edge cases—such as emergency collision avoidance maneuvers or navigating low-friction surfaces—is paramount for developing truly robust and safe autonomous systems capable of handling unexpected events beyond nominal driving conditions \cite{zhu_control_2024}.

% Context of Racing and Electrification/A4WD added
Autonomous racing serves as an ideal, albeit extreme, testbed for developing and validating control strategies under such demanding conditions \cite{betz_autonomous_2022}. In a racing environment, vehicles are consistently pushed to their physical limits, requiring controllers to exploit the maximum available tire grip to achieve optimal performance, typically measured in minimum lap time. Algorithms proven effective in this high-stakes domain can offer valuable insights and technologies applicable to enhancing the safety and performance envelope of conventional autonomous road vehicles. Concurrently, the automotive industry is undergoing a significant change. Alternative drivetrains gain traction, and shift towards electrification is more and more widespread. Electric vehicles (EVs), particularly those equipped with \gls{a4wd} systems featuring independently controllable motors for each wheel, are becoming increasingly prevalent, especially in the performance sector. This powertrain architecture offers unprecedented control authority, enabling rapid and precise torque distribution to individual wheels. Such capabilities unlock advanced \gls{vdc} possibilities like torque vectoring, which can significantly enhance vehicle handling, stability, and agility, especially when operating near the limits of adhesion.

% Challenges of A4WD control and limitations of classical methods expanded
However, fully harnessing the potential of \gls{a4wd} presents considerable control challenges. The increased number of actuators adds complexity, requiring sophisticated coordination strategies to manage the intricate interplay between longitudinal and lateral forces at each tire \cite{sakai_motion_1999, yoichi_hori_future_2004}. Accurately modeling the highly nonlinear tire behavior under combined slip conditions, along with the vehicle's complex dynamics at high speeds, is difficult yet crucial for traditional model-based control design \cite{yu_lateral-stability-oriented_2024}. Classical control approaches, including robust control \cite{he_optimum_2006, xiong_research_2011} and \gls{mpc} \cite{max_time_2014, zhao_model_2015, kabzan_learning-based_2019}, have been extensively researched for \gls{a4wd} control. While effective within certain operating ranges, they often rely on simplified models, require meticulous parameter tuning, and can struggle to adapt optimally across the full spectrum of dynamic conditions encountered at the grip limit, potentially leading to conservative or suboptimal performance.

% Potential of RL/DRL introduced and explained
Recent advancements in \gls{ai}, particularly in the field of \gls{drl}, offer a compelling alternative paradigm for tackling these complex control problems \cite{goodfellow_deep_2016, sutton_reinforcement_2018}. \gls{drl} algorithms have demonstrated remarkable success in learning complex policies for decision-making and control directly from interaction data, often achieving superhuman performance in domains ranging from computer games \cite{openai_dota_2019, vinyals_grandmaster_2019, mnih_human-level_2015} to complex robotic manipulation \cite{openai_solving_2019} and locomotion \cite{heess_emergence_2017, haarnoja_soft_2018}. The key advantage of \gls{rl} lies in its potential to automatically discover optimal control strategies for high-dimensional, nonlinear systems without relying on explicit, predefined models. By optimizing a policy based on cumulative rewards related to the task objective (e.g., maximizing track progress or minimizing lap time), \gls{rl} agents can potentially learn non-intuitive yet highly effective control behaviors that exploit the full capabilities of the underlying hardware.

% Introduce specific work, contributions, and structure
This work investigates the application of \gls{drl} to the specific challenge of controlling an \gls{a4wd} racecar for time-optimal performance. We employ the \gls{ppo} \cite{schulman_proximal_2017} algorithm, a robust and widely used on-policy \gls{rl} method, to train a single \gls{nn} policy. Distinctively, this agent learns to map vehicle dynamic states (such as velocities, yaw rate, accelerations) directly to low-level actuator commands: a steering angle and, critically, \textit{independent} torque commands for each of the four wheels. This end-to-end formulation bypasses the need for intermediate controllers or explicit torque allocation modules. The agent must implicitly learn how to best distribute torque among the wheels to maximize progress while maintaining stability and adhering to track limits, effectively solving the complex \gls{a4wd} control problem as an integrated part of the driving policy. We hypothesize that this approach allows the agent to discover sophisticated torque vectoring strategies tailored to limit-handling performance in racing scenarios. The resulting policy is evaluated in the high-fidelity racing simulator TORCS \cite{noauthor_torcs_2020}. Through detailed analysis, including corner-by-corner examination of control actions and resulting vehicle behavior (e.g., GG-diagrams), we demonstrate the effectiveness of the learned policy and compare its performance against traditional approaches. This study aims to highlight the potential of end-to-end \gls{rl} for developing unified, high-performance control systems for vehicles with complex actuation, offering insights relevant to both autonomous racing and the broader field of advanced vehicle dynamics control.

The remainder of this paper is organized as follows: Section \ref{sec:methods} details the problem formulation and the specifics of our \gls{rl} approach, including the environment, agent architecture, and reward design. Section \ref{sec:results} presents simulation results, analyzing the learned driving behavior and comparing performance. Finally, Section \ref{sec:conclusion} concludes the paper and discusses potential future work.

%========================================================================
\section{Related Work}
\label{sec:relatedwork}

The control of vehicles at their handling limits, particularly for \gls{a4wd} systems, has been approached through various methodologies, ranging from classical control to modern reinforcement learning. Understanding these approaches helps position the contributions of the present work.

\subsection{Classical and Learning-Enhanced Control Methods}
Classical control techniques, such as robust control \cite{he_optimum_2006, xiong_research_2011} and \gls{mpc} \cite{max_time_2014, zhao_model_2015}, have been extensively studied for \gls{vdc}. Robust control methods prioritize guaranteed stability margins, often crucial for passenger vehicle safety, while \gls{mpc} excels at optimizing performance within known constraints, making it popular for racing applications. For example, \cite{yu_lateral-stability-oriented_2024} utilized \gls{mpc} with dynamically updated tire force constraints to enhance path tracking stability in extreme, but relatively simple, maneuvers (e.g., double lane change) simulated in CarSim, contrasting with our work's focus on full-lap time optimization using RL. \cite{de_buck_minimum_2023} specifically tackled minimum lap time optimization for \gls{a4wd} using \gls{nlp} via direct collocation, optimizing torque vectoring and active aerodynamics. While similar to our work in its racing focus and lap time evaluation, it employed a simpler 3DOF vehicle model and classical optimization rather than an end-to-end RL approach.

Recognizing the limitations of purely physics-based models, especially under varying conditions or at performance limits, researchers have integrated learning into classical frameworks. \cite{kabzan_learning-based_2019} demonstrated a learning-based \gls{mpc} on a full-size autonomous (Fomrula Student) race car. They used Gaussian Processes (GPs) trained online from vehicle state measurements (position, orientation, velocities, yaw rate, steering angle, throttle/brake command) to estimate and compensate for the error of a nominal dynamic bicycle model within the \gls{mpc}. This approach differs from our work which learns the entire control policy via RL rather than learning model corrections for \gls{mpc}. Similarly, \cite{frohlich_contextual_2022} focused on adapting \gls{mpc} tuning parameters (cost weights) for autonomous racing on 1:28 scale RC cars under changing environmental conditions (emulated by varying friction). They used a learned parametric dynamics model to infer a low-dimensional 'context' vector. This context was then used within contextual Bayesian Optimization to efficiently find optimal \gls{mpc} tuning parameters for different conditions, evaluated by lap time on the physical platform. This differs from our approach where the RL agent implicitly adapts to the environment based on state observations, without explicit context variables or separate tuning optimization. \cite{lu_two-timescale_2023} proposed a hybrid mechanism-and-data-driven controller for aggressive driving in TORCS, explicitly separating the control problem based on timescales. A mechanism-based \gls{mpc} controller operated on the slow timescale (car body pose/velocity) to determine desired friction forces, while a data-driven module (estimating tyre parameters) helped translate these forces into fast-timescale wheel torques and steering angles. This modular, timescale-aware design contrasts with our single-policy end-to-end RL method but shares the goal of fusing mechanism and data for improved performance and adaptability.

\subsection{Reinforcement Learning Approaches}
Driven by successes in complex tasks \cite{mnih_human-level_2015}, \gls{rl} offers an alternative paradigm, learning control policies directly from interaction. However, transferring RL policies trained in simulation to the real world remains challenging \cite{jiang_sim--real_2022, tsounis_deepgait_2020}. Applications in the automotive domain include energy management \cite{zhiyan_feng_practicability_2023} and active suspension control \cite{nhu_physics-guided_2024}, where \gls{rl} learned control actions (actuator stiffness/damping) based on vehicle state inputs, guided by physics-based constraints and rewards related to passenger comfort and stability.

Applying \gls{rl} to low-level \gls{vdc} like \gls{a4wd} is emerging. \cite{wei_deep_2022} used TD3-\gls{ddpg} in simulation (double lane change) to learn corrective torque distributions, modifying a reference torque based on steering angle, using vehicle dynamic states as input and focusing on stability and energy saving. \cite{deng_deep_2023} employed a hierarchical structure in simulation: a safety layer defined target moments/forces, and a \gls{drl} agent (using speed, yaw deviation, etc. as state) performed torque allocation (action) for efficiency and safety. \cite{hua_path_2024} used an \gls{rl}-based auxiliary controller within a larger framework for path tracking in four-wheel steer/drive vehicles in simulation, focusing on improving sample efficiency with a GER mechanism for simpler scenarios. These differ from our work's integrated, end-to-end approach where a single policy directly outputs both steering and individual wheel torques.

In autonomous racing, \gls{rl} has shown significant promise. \cite{kaufmann_champion-level_2023} achieved superhuman performance in real-world drone racing using \gls{drl} trained in simulation combined with real-world data. \cite{czechmanowski_learning_2025} successfully trained an \gls{rl} policy (\gls{ppo}) using domain randomization and actuator modeling in simulation that achieved zero-shot transfer to a real F1TENTH car, outperforming both \gls{mpc} and an expert human RC driver. Their agent used geometric track features, vehicle state (velocities, yaw rate, steering angle, previous actions) as input, outputting steering and wheel speed commands. The reward was based on track progress, similar to our primary reward component. Key differences from our work include the platform (F1TENTH RC car vs. simulated racecar), the action space (steering/speed vs. steering/individual torques), and the real-world validation. \cite{kalaria_towards_2023} specifically addressed head-to-head racing using \gls{rl} in a custom dynamic vehicle simulation. They proposed a curriculum learning framework transitioning from simple to complex models, with rewards related to checkpoint progress and opponent interaction. Their focus on competitive scenarios and curriculum learning differs from our time-trial focus. \cite{subosits_towards_2024} developed an \gls{rl} agent (DSAC) functioning as an autonomous test driver in a high-fidelity simulation (Lexus RC F GT3 in Dymola). The agent observed vehicle state (velocities, accelerations, inputs), track geometry (lookahead points), and crucially, vehicle setup parameters, outputting steering and throttle/brake commands. Trained in a multi-task setting across different setups (varying grip/balance), it matched/exceeded human lap times and correctly predicted performance trends. They also incorporated imitation learning (using human demonstrations in the replay buffer and policy loss) to encourage human-like driving styles, potentially altering the reward to penalize path deviation from human lines. This work shares the \gls{rl} methodology and high-performance goal but focuses on setup evaluation and imitation, unlike our direct A4WD control learning. The simulation environment is also different (Dymola vs. TORCS). Simulation tools like AARK \cite{bockman_aark_2024}, based on Assetto Corsa, aim to facilitate such research by providing open interfaces and data generation capabilities for high-fidelity simulators.

This work builds upon these foundations by employing an end-to-end \gls{rl} approach (\gls{ppo}) where a single policy maps vehicle dynamic states (velocities, yaw rate, accelerations, etc.) directly to steering commands and independent wheel torques. This allows the agent to implicitly learn complex \gls{a4wd} control strategies, including stability functions like \gls{abs} and \gls{asr}, within the driving policy to achieve time-optimal performance on a simulated racetrack (TORCS environment). The primary reward is based on maximizing progress along the track centerline, augmented with penalties for constraint violations. Compared to our precursor work \cite{bari_vision_2025}, the key differences are the direct control of wheel torques instead of pedal abstractions and the use of vehicle dynamic states as input instead of raw visual data. Our approach contrasts with hierarchical methods, those focusing solely on trajectory following or simpler maneuvers, or those relying on classical optimization enhanced by learning.

%======================================================================
\section{Methods}
\label{sec:methods}

\begin{figure}[!h]
 \centering
 \includegraphics[width=8.0cm]{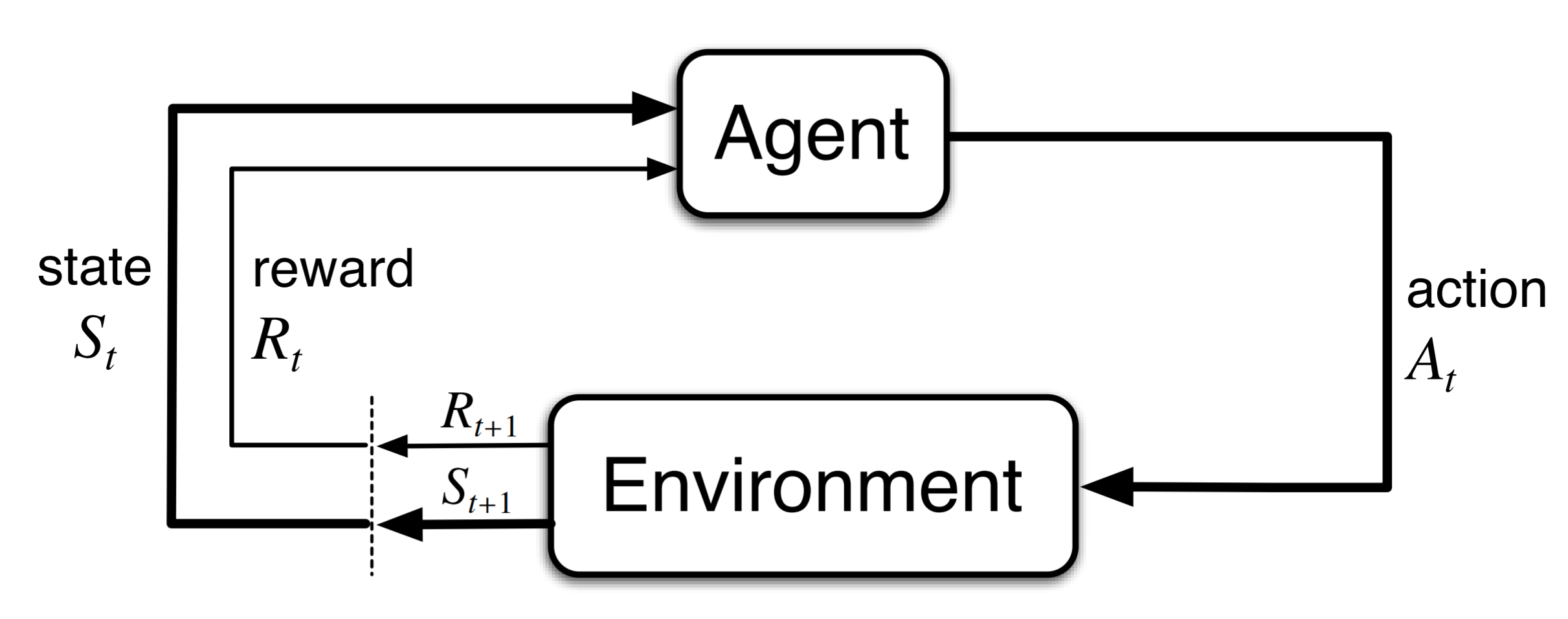} % Adjusted size slightly
 \caption{Scheme of the Markov Decision Process, formalizing the Agent-Environment interaction in \gls{rl} problems \cite{sutton_reinforcement_2018}.} % Used \gls
 \label{fig:RL}
\end{figure}

This work addresses the \gls{a4wd} control problem by framing the task of time-optimal race car driving as a \gls{rl} problem. This approach allows a single trained agent to handle path following, stabilization, and low-level torque vectoring in an integrated manner. Racing provides a suitable environment for this study, as vehicles operate consistently at the tire grip limit, highlighting the benefits and challenges of advanced drivetrain control. We follow the standard \gls{rl} framework, often formalized as a Markov Decision Process (Figure~\ref{fig:RL}), involving an agent interacting with an environment \cite{sutton_reinforcement_2018}. For clarity, Table~\ref{tab:CERLterms} maps common \gls{rl} terms to their analogues in control engineering; note also that 'driver modeling' and 'vehicle control' can be considered analogous tasks in the context of developing autonomous driving algorithms. The \gls{rl} problem was implemented using the Stable Baselines3 package \cite{raffin_stable-baselines3_2021}.

\begin{table}[h]
\tbl{Analogous terms in Control Engineering and in Reinforcement Learning.} % Shortened caption
{\begin{tabular}{ll} 
 \toprule
  Reinforcement Learning & Control Engineering \\  
 \midrule
 Agent & Controller \\ 
 Environment & Plant, System \\
 Action & Control signal, System input or Controller output \\
 State, Observation & Measurement, System output or Controller input \\ 
 \bottomrule
 \end{tabular}}
 \label{tab:CERLterms}
\end{table} 

\subsection{Environment and Vehicle Models} % Added subsection for clarity
\begin{figure}[!h]
 \centering
 \includegraphics[width=12.5cm]{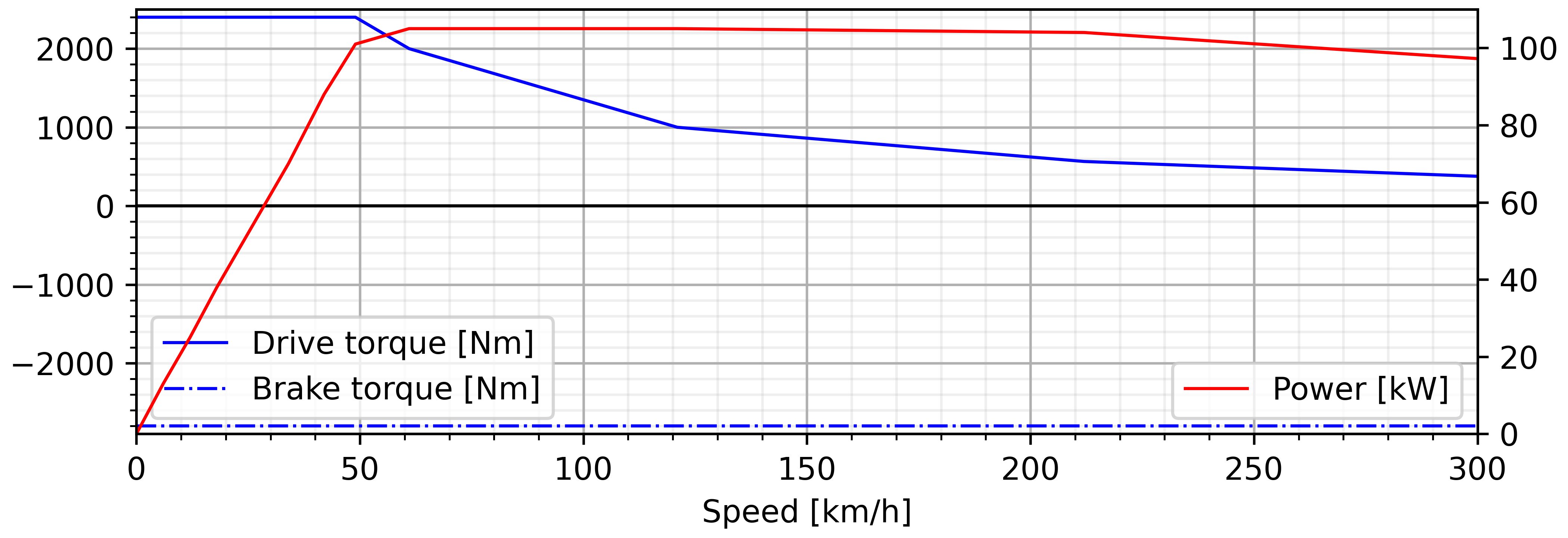}
 \caption{Characteristics used to model wheel motors for the active \gls{4wd} car.} % Used \gls
 \label{fig:torque}
\end{figure}

The \textbf{Environment} used is the open-source car racing simulator TORCS \cite{noauthor_torcs_2020}, known for its adaptability for \gls{ai} research and its provision of detailed vehicle dynamic signals. One racetrack ('brondehach', see Table~\ref{tab:hyper}) was used. Two vehicle models were employed: 
\begin{enumerate}
    \item A built-in TORCS 4WD car model representing a conventional 'passive \gls{4wd}' system, controlled via standard pedal and steering inputs (used for baseline comparison later)
    \item A modified version, the 'active \gls{4wd}' car, controlled via steering and four independent wheel torques. This car's powertrain was modeled using simple electric motor characteristics for each wheel (Figure~\ref{fig:torque}). The combined maximum power of these four motors was set equal to the maximum engine power of the passive \gls{4wd} car to focus the comparison on control strategy differences, although inherent electric motor characteristics may still offer performance advantages, as discussed in the Results section.
    
\end{enumerate}

\subsection{Observation and Action Space} % Added subsection for clarity
The \textbf{Observation} provided to the agent consists of a vector of vehicle dynamic states, detailed in Table~\ref{tab:observ}. The \textbf{Action} space for the active \gls{4wd} agent comprises five continuous values normalized to the range [-1, 1]: one for the \gls{swa} (conventional front steering, +1 max left, -1 max right) and one for each of the four wheel torques (+1 max driving torque, -1 max braking torque). Agent-environment interaction occurred at 20Hz. Notably, no explicit rate limits were applied to the actions, allowing the agent maximum control authority without human-imposed constraints. This 20Hz frequency, however, has consequences compared to typical human or \gls{vdc} control bandwidths: steering inputs can appear significantly more abrupt ('harsh') than those of a human driver (~10Hz), while the torque control operates much slower than typical \gls{vdc} systems (often >=100Hz). The relatively slow torque updates imply that potentially higher performance could be achievable with increased control frequency, especially for torque modulation.

\subsection{Reward Function} % Added subsection for clarity
The \textbf{Reward} signal guides the agent's learning towards maximizing performance, defined here as minimizing lap time. To provide dense feedback crucial for efficient learning (avoiding sparse rewards like raw lap time), we primarily used a 'progress reward' \cite{fuchs_super-human_2020, wurman_outracing_2022}. This reward is the distance traveled along the track centerline ($s^{cl}$) in the last timestep ($t$), shown in Eq.~\eqref{eq1}, which approximates the time difference information used by human racers.

\begin{equation} \label{eq1}
r^{progr.}_{t} = s^{cl}_{t} - s^{cl}_{t-1}
\end{equation}

where $s^{cl}_{t}$ is the distance traveled along the track centerline until timestep $t$. 
To improve training robustness, additional components were added. A termination reward ($r^{ter}_{t}$) provided a large positive value for completing a lap and penalties for terminating early due to various reasons (e.g., off-track, damage, low progress). An action penalty ($r^{act}_{t}$), defined in Eq.~\eqref{eq:actbound}, discouraged the agent from outputting actions outside the feasible [-1, 1] range, which stabilized training.

\begin{equation}
  \label{eq:actbound}
  r^{act}_{t}= (\frac{|a_{t}|}{p^{sc}} - p^{bnd} + 1)^{2} 
\end{equation}

Here, $a_t$ represents the action vector at time $t$, and $p^{sc}$ and $p^{bnd}$ are scaling and boundary parameters (see Table~\ref{tab:reward}). The final reward function is the sum of these components, as shown in Eq.~\eqref{eq2}:

\begin{equation} \label{eq2} % Note: Original was Eq 3, adjusted for flow
r_t = r^{progr.}_{t} + r^{ter}_{t} - r^{act}_{t}
\end{equation}
Specific parameter values for the reward function are detailed in Table~\ref{tab:reward}.

\subsection{Agent Architecture and Training} % Added subsection for clarity
The \textbf{Agent} is implemented as a single \gls{nn} using an \gls{mlp} architecture. The network features three shared hidden layers (300, 600, 600 neurons) for both the policy and value functions, plus an additional layer (600 neurons) for the value function only, all using ReLU activation functions. 
The \gls{ppo} algorithm \cite{schulman_proximal_2017} was selected for training due to its demonstrated robustness and effectiveness in various complex tasks. Training utilized multiple (32) TORCS instances running in parallel to gather experience efficiently. Trainings were conducted on DELL R730 hardware and capped at approximately $10^9$ total steps (around 10 days). Agent performance was evaluated every 10,000 steps in 'test mode', where the deterministic mean of the policy network's output distribution was used for actions, contrasting with 'training mode' where actions were sampled stochastically. Training hyperparameters are listed in Table~\ref{tab:hyper}.

\section{Results}
\label{sec:results}

\begin{figure}[h]
\centering
\resizebox*{14cm}{!}{\includegraphics{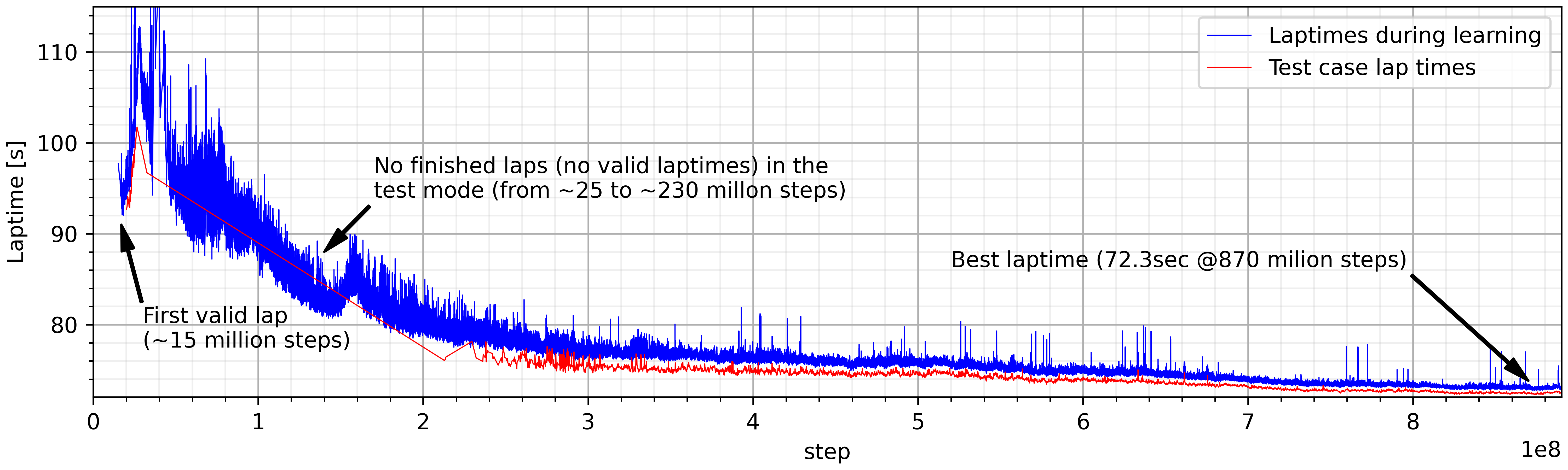}}
\caption{Learning curve showing lap time improvement during training. 'Explored Laptimes' correspond to training episodes with stochastic actions, while 'Exploited Laptimes' correspond to test laps using deterministic actions (mean of policy output).} 
 \label{fig:Learning}
\end{figure}

Figure~\ref{fig:Learning} shows lap times as a function of the training steps. The blue curve represents the lap times achieved during training episodes where actions were sampled stochastically, while the red curve shows the lap times during evaluation episodes (test laps) using deterministic actions (the mean of the policy network's output distribution). An interesting observation is that while the agent learns to navigate the track relatively quickly (completing laps after $\sim$15 million steps), it requires significantly more training ($\sim$230 million steps) to consistently complete laps in the deterministic test mode. Furthermore, substantial lap time improvement occurred much later, with an additional $\sim$600 million steps needed to gain approximately 2 seconds. Figure~\ref{fig:fulllap} illustrates a complete lap driven by the agent in test mode after approximately 870 million training steps.

\begin{figure}[!h]
\centering
\resizebox*{14cm}{!}{\includegraphics{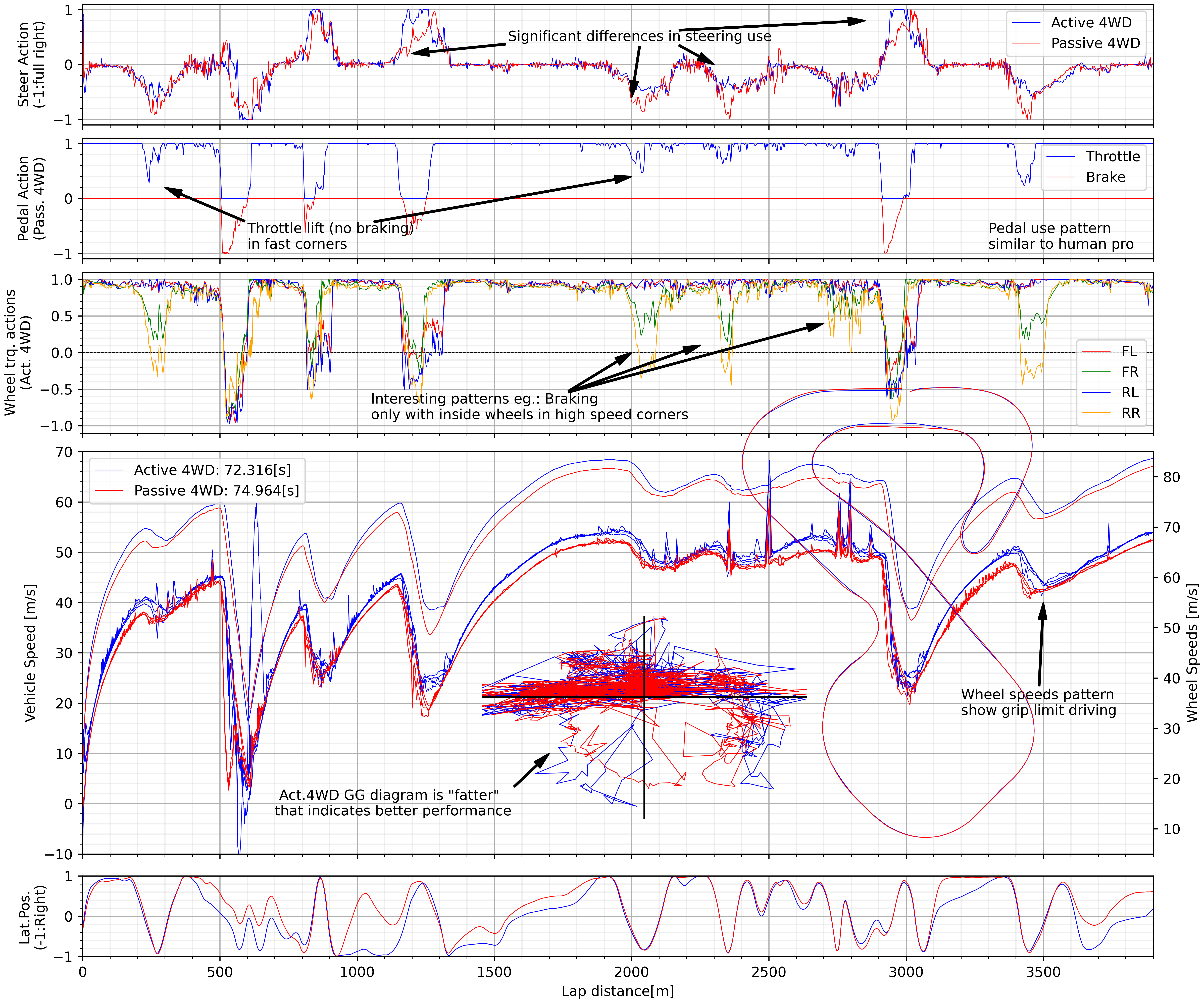}}
\caption{Comparison of representative laps between the agent controlling the passive \gls{4wd} vehicle (via steering/pedals) and the agent controlling the active \gls{a4wd} vehicle (via steering/individual torques). Plots show (top to bottom): steering wheel angle, throttle/brake pedal (passive only), individual wheel torques (active only), vehicle and wheel speeds, and lateral track position. The central plot shows the GG-diagram (lateral vs. longitudinal acceleration) for both agents.} 
 \label{fig:mech_vs_AI}
\end{figure}

Figure~\ref{fig:mech_vs_AI} provides a detailed comparison between two representative laps: one driven by an agent trained for a standard 'passive \gls{4wd}' vehicle model in TORCS (controlled via steering and pedals), and the other driven by the primary agent of this study controlling the 'active \gls{a4wd}' vehicle via steering and independent wheel torques. Both laps represent the best performance achieved after approximately $10^9$ training steps. The horizontal axis in these plots represents the distance traveled along the track centerline.

The top graph, showing steering traces for both vehicles (-1: max right, +1: max left), reveals significant differences, indicating how individual wheel torque control enables distinct driving strategies compared to the passive system. The second graph (throttle/brake pedal input, +1: max throttle, -1: max brake) applies only to the passive \gls{4wd} agent and displays patterns typical of race driving: full throttle on straights and sharp braking before corners. The third graph illustrates the four independent wheel torque commands (actions) for the active \gls{a4wd} agent (+1 max driving, -1 max braking). This plot reveals complex and dynamic torque modulation: torques differ between wheels, vary across different corners, and change even within different phases of a single corner, showcasing the learned control strategy that will be analyzed further below.

The fourth graph displays vehicle velocity and individual wheel speeds. A clear observation is the superior straight-line acceleration of the active \gls{a4wd} vehicle. This difference primarily stems from the powertrain modeling choices described in Section~\ref{sec:methods}: the active \gls{a4wd} car uses simulated electric motors at each wheel. While the 'combined peak power' was matched to the passive car's engine, the inherent torque characteristics of electric motors allow for better power utilization across a wider speed range, leading to faster acceleration. This performance advantage, while contributing significantly to lap time reduction, is partially due to the powertrain difference itself, not solely the learned \gls{rl} control strategy. Therefore, subsequent analysis will focus more on handling behavior and control patterns rather than solely on overall lap time to assess the effectiveness of the learned torque vectoring. This distinction is important for interpreting the results accurately.

The final graph shows the lateral track position (+1 left limit, -1 right limit). Notably, despite the agents being trained independently on vehicles with fundamentally different actuation, they converged on very similar optimal racing lines.

Embedded within the speed trace graph is the 'GG-diagram', plotting the vehicle's longitudinal versus lateral acceleration throughout the lap. A wider, more rounded GG-diagram generally indicates driving closer to the vehicle's limits, while a thinner, cross-like shape suggests less aggressive driving. The GG-diagram comparison reveals a crucial insight into the active \gls{a4wd} agent's effectiveness: it consistently achieves higher combined accelerations, particularly during the combined braking and turn-in phases of corners. Quantitatively, the active \gls{a4wd} agent reaches approximately 25\% higher combined g-forces in these critical phases compared to the passive \gls{4wd} agent. This enhancement is less pronounced in pure lateral acceleration or pure longitudinal acceleration/braking, logically highlighting the primary benefit of active torque control: the ability to precisely distribute torque to each wheel allows the agent to more efficiently utilize the individual grip potential available at each tire during phases where forces are combined, a capability far exceeding the simpler torque distribution of a mechanical \gls{4wd} system or conventional braking.

It is also noticeable from the steering trace and GG-diagram that the baseline passive \gls{4wd} car exhibits significant \gls{us} behavior, frequently reaching maximum steering lock in corners. This characteristic is common in simulators like TORCS, where \gls{us} setups provide stability that compensates for the reduced sensory feedback available to human drivers compared to real-world driving. Speed and distance are more difficult to estimate on the screen compared to real life, and vibrations, accelerations, and forces are not felt in simulation. While professional sim racers often customize their own setups, the default TORCS setup was used here. The active \gls{a4wd} agent must therefore learn to overcome this inherent \gls{us}.

\begin{figure}[!h]
\centering
\resizebox*{14cm}{!}{\includegraphics{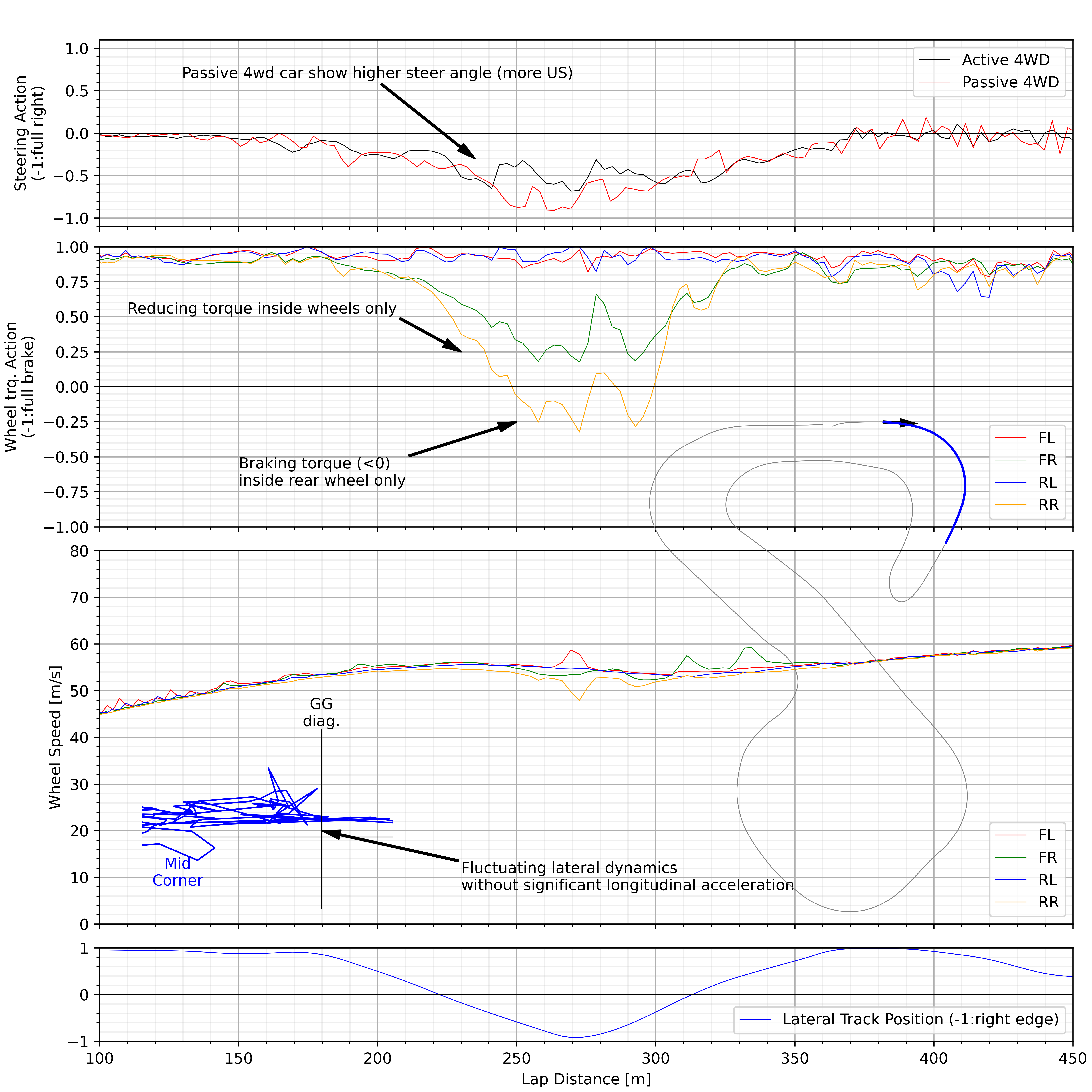}}
\caption{Detailed driving behavior of the active \gls{a4wd} agent in Turn 1. Includes steering comparison with the passive \gls{4wd} agent for reference.} 
 \label{fig:4wdzoomT1.png}
\end{figure}

To gain deeper insight into the learned \gls{a4wd} control strategy, we analyze vehicle behavior in representative turns. The turns on the 'brondehach' track are categorized into three groups based on their characteristics]. Since behavior within each group is similar, one turn per category is analyzed in detail.

The first group (Turns 1, 9, and sequence 5-6-7) consists of fast, right-hand corners typically navigated near full throttle, primarily representing the 'Mid-Corner' phase with minimal longitudinal acceleration. Figure~\ref{fig:4wdzoomT1.png} shows Turn 1 as an example (Turns 5-6-7 and 9 are shown in Appendix Figures~\ref{fig:4wdzoomT567} and \ref{fig:4wdlast}). The steering comparison (top graph) is informative: the active \gls{a4wd} agent uses significantly less steering angle mid-corner ($\sim$270m) than the passive agent, indicating reduced \gls{us}. The torque plot (second graph) reveals the reason: approaching the corner ($\sim$180m), the agent reduces driving torque, but primarily on the inside wheels (Front Right and Rear Right). Applying less torque to the inside wheels, and even negative (braking) torque to the inside rear wheel, generates a yaw moment that counteracts the base vehicle's \gls{us} tendency. The lateral position (bottom graph) shows a typical racing line, utilizing the full track width.

\begin{figure}[!h]
\centering
\resizebox*{14cm}{!}{\includegraphics{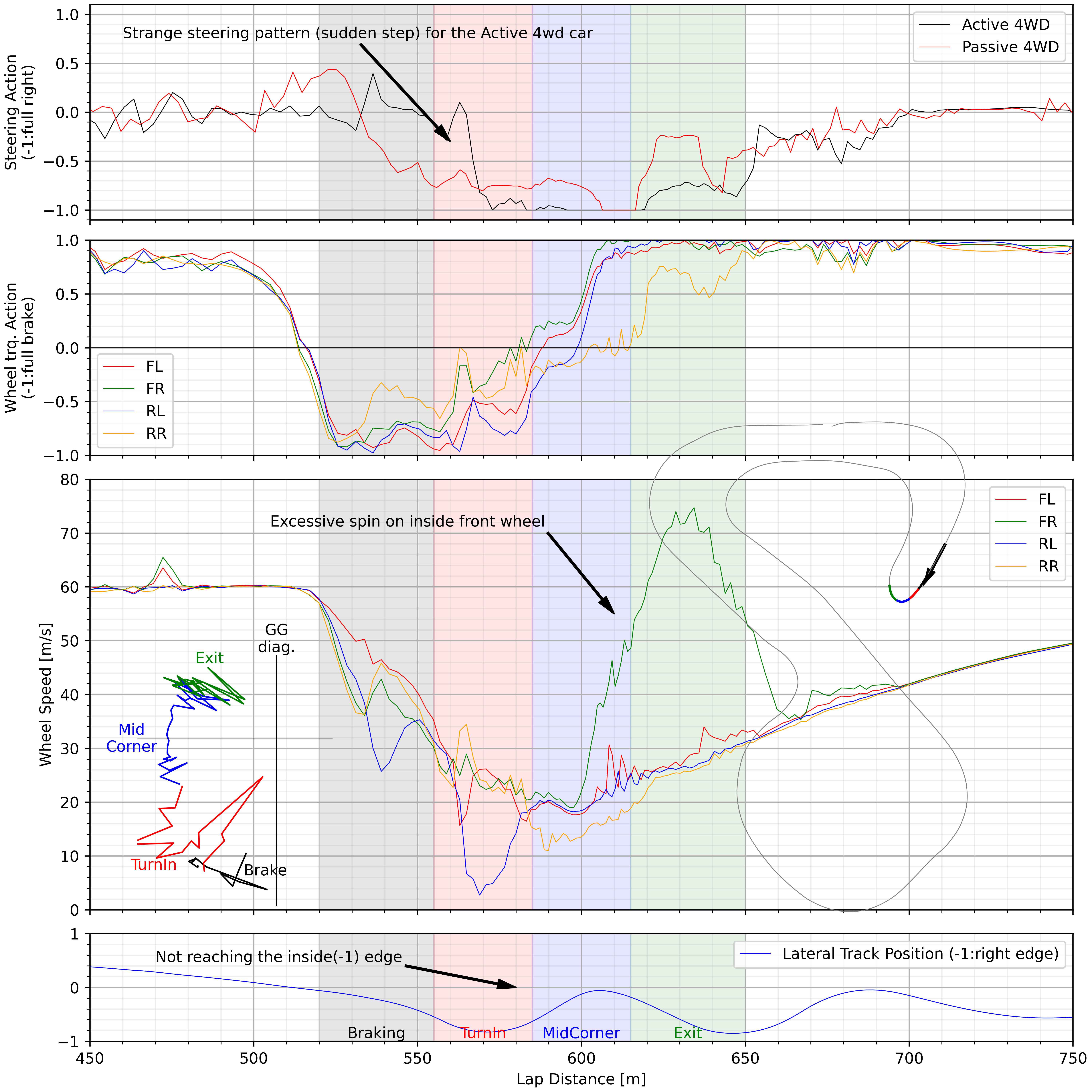}}
\caption{Detailed driving behavior of the active \gls{a4wd} agent in Turn 2 (slow hairpin).} 
 \label{fig:4wdzoomT2}
\end{figure}

Figure~\ref{fig:4wdzoomT2} illustrates Turn 2, a slow, tight 'hairpin' corner ($\sim$180° turn) which showcases all cornering phases (braking, turn-in, mid-corner, exit)]. Several unusual patterns emerge here. The steering trace shows a sudden application to maximum lock near the end of straight-line braking ($\sim$520m). The wheel speed plots indicate excessive slip on the inside front wheel (FR) during corner exit, and the lateral position trace reveals the agent does not reach the geometric apex (inside edge) of the corner.

This seemingly counter-intuitive style could be either suboptimal learning or an optimal strategy adapted to the simulation's specific characteristics. While a definitive answer requires further study, several factors within TORCS support the latter possibility. Tire modeling in TORCS is relatively simple, not accurately capturing behavior at extreme slip (like locked wheels or excessive spin) nor effects like overheating or wear. Human drivers report the inside of Turn 2 as bumpy, potentially compromising grip and exit acceleration. Finally, the short straight following Turn 2 influences the optimal exit line.

Given this context, the agent's strategy might be optimal within TORCS. The yaw rotation instability during braking is controlled partly by differential torque rather than solely steering. This is a known \gls{vdc} technique which is used by the agent potentially to exploit the tire model's friction characteristics. After an unstable TurIn phase, in mid-corner cars thends to \gls{us} (under-rotate). In this case, applying maximum steering mid-corner  indicates \gls{us}, and induces high lateral slip on the front tires, possibly pushing them beyond their accurately modeled range. Similarly, the excessive spin on the inside front wheel (FR), while potentially detrimental in reality (due to heat/wear), incurs no penalty in TORCS's simplified model and might be exploited by the agent if it aids rotation or acceleration in that specific state. Analysing wheel speed vs. torque across different corners reveals learned behaviors mimicking \gls{abs} (preventing lock-up during braking) and \gls{asr} (managing spin during acceleration), demonstrating the agent integrates stability functions implicitly. The inside rear (RR) torque also lags in reaching its maximum, further aiding yaw rotation. The choice to avoid the bumpy apex and position the car for the subsequent short straight by not running wide on exit aligns with lap time optimization under these specific simulated constraints. 

\begin{figure}[!h]
\centering
\resizebox*{14cm}{!}{\includegraphics{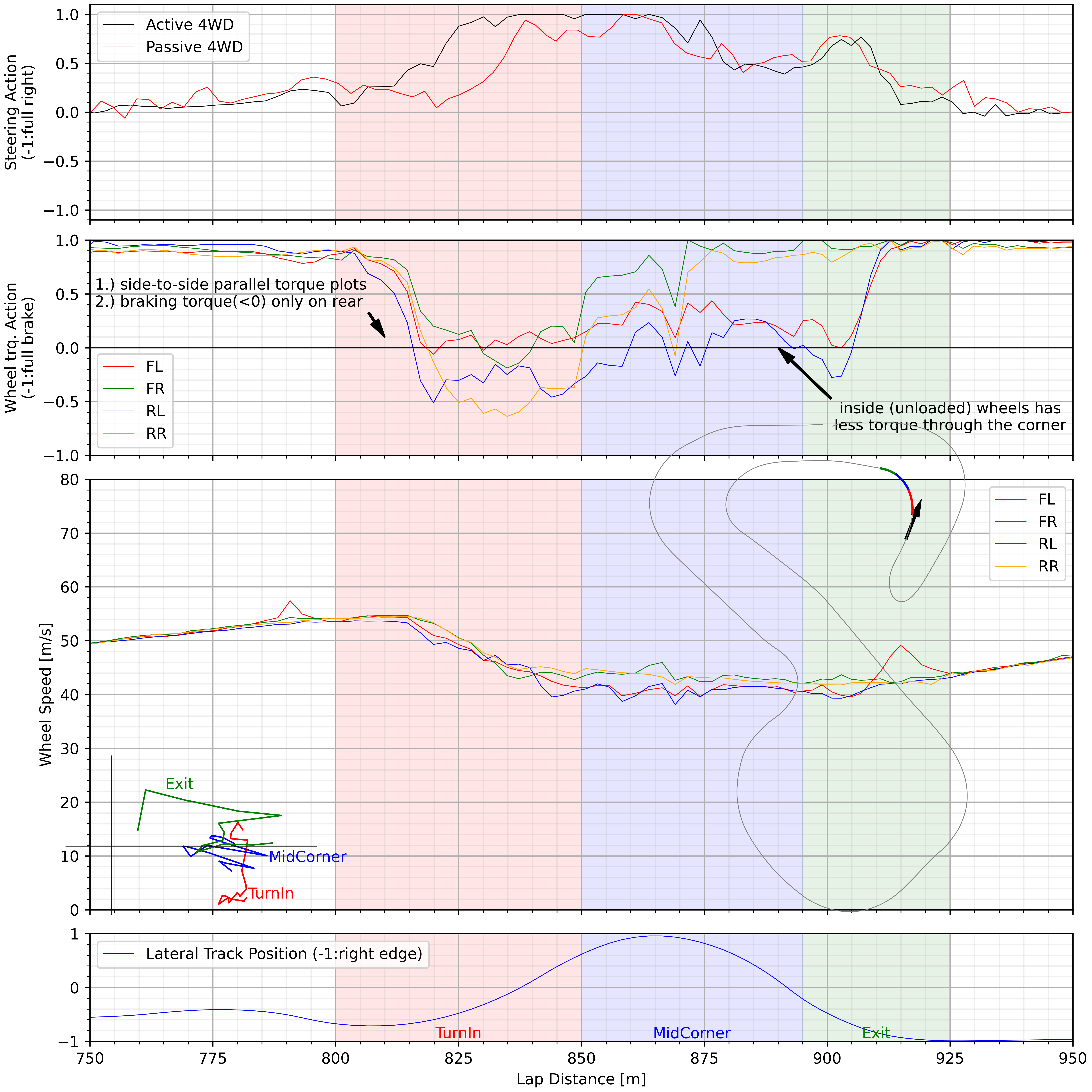}}
\caption{Detailed driving behavior of the active \gls{a4wd} agent in Turn 3.} 
 \label{fig:4wdzoomT3}
\end{figure}

Figure~\ref{fig:4wdzoomT3} shows Turn 3, representative of the third group (Turns 3, 4, 8) characterized by smaller speed changes and braking occurring under lateral load (turn-in phase dominant). Similar to Turn 2, the steering trace shows high angles are used. The wheel torque plots show torques on the same side (FL/RL and FR/RR) running roughly in parallel, suggesting left-right torque differences are again used for yaw rotation stability. Notably, similar to the high-speed corners (group 1), braking torque is applied primarily to the rear wheels, while front wheel torques remain near zero. Applying braking force mainly at the rear reduces stabilizing rear lateral grip, while high steering angles with minimal longitudinal force at the front maximize front lateral grip, a strategy that counteracts \gls{us}.

A consistent pattern observed across corners is that during both braking and acceleration, the outside wheels experience larger magnitude torques than the inside wheels. This aligns with physics: lateral weight transfer increases normal load on the outside tires, allowing them to generate higher longitudinal forces. The agent exploits this by demanding more braking force from the outer wheels on entry and more acceleration from them on exit. However, this optimal force utilization inherently induces a yaw moment: the side-to-side difference in longitudinal braking forces tends to cause \gls{us} on entry, while the difference during acceleration (exit) tends towards \gls{os}. The agent's learned control patterns reflect a continuous management of this corner-phase-dependent \gls{us}/\gls{os} shift, overlaid on the vehicle's base (game-like) \gls{us} characteristic. (Remaining turns in this group, 4 and 8, are shown in Appendix Figures~\ref{fig:4wdzoomT4} and \ref{fig:4wd2nd2last}).

\section{Conclusion}
\label{sec:conclusion}

This work successfully applied \gls{rl}, specifically the \gls{ppo} algorithm, to the challenging domain of high-performance autonomous driving with \gls{a4wd}. We demonstrated that an agent can learn effective control policies by directly mapping vehicle states to steering angle and, crucially, independent wheel torque commands for each of the four wheels. This approach bypasses the need for separate, pre-designed controllers for subsystems like torque vectoring. Analysis of the learned behaviors revealed sophisticated strategies emerging directly from the learning process, such as proactive torque adjustments based on cornering phase and predicted load transfer, effectively mitigating the base vehicle's inherent \gls{us} characteristics (as observed in Turns 1 and 3, for example). Furthermore, the agent implicitly developed functionalities analogous to classical \gls{vdc} systems like \gls{abs} and \gls{asr} by modulating torques to maintain traction, and achieved active yaw control through differential torque application, all without explicit rules for these functions being encoded. The policy also showed adaptation to the specific constraints and nuances of the simulation environment, including the tire model characteristics.

The performance achieved by the \gls{rl} agent resulted in lap times that significantly surpassed those of a comparable passively controlled \gls{4wd} vehicle operating within the same simulation environment. While direct lap time comparisons are partially influenced by the differing acceleration profiles inherent to the modeled electric versus conventional powertrains, the detailed analysis of control inputs and vehicle response (e.g., through GG-diagrams and corner-specific torque patterns) clearly indicates substantial improvements in handling dynamics and stability attributable to the learned active control strategy. This data-driven optimization demonstrates \gls{rl}'s potential to uncover non-intuitive yet effective control laws for highly coupled, nonlinear systems like \gls{a4wd} vehicles, offering a promising avenue compared to traditional physics-based methods which may struggle with such complex interactions or require extensive manual tuning. The ability of the agent to learn these integrated strategies highlights a key advantage of end-to-end learning approaches for complex control problems.

Several promising directions for future research emerge from this work. Evaluating the agent's generalization capabilities across a wider variety of racetracks with different characteristics would be crucial for assessing the robustness and adaptability of the learned policies. Extending the agent's action space to include control over other actuators, such as \gls{as}, presents an exciting opportunity to explore fully integrated vehicle dynamics control through \gls{rl}, potentially leading to further performance gains by coordinating suspension and drivetrain actions. Perhaps most significantly, bridging the simulation-to-reality gap by testing and adapting these learned policies on a physical vehicle platform remains a critical next step towards real-world applicability.

In conclusion, this study provides compelling evidence for the capability of deep \gls{rl} to generate sophisticated, unified control strategies for complex vehicle dynamics problems like \gls{a4wd} racing. By demonstrating how an agent can learn near-optimal control directly from interaction and reward signals, this work contributes to the growing field of autonomous racing and pushes the boundaries of \gls{ai} application in high-performance scenarios. The insights gained from controlling vehicles at their operational limits in demanding environments like racetracks, particularly regarding the coordinated control of multiple actuators, can ultimately inform the development of more adaptive, robust, and potentially safer autonomous driving systems for general road use, contributing to the broader goals of enhanced vehicle safety and efficiency.

\bibliographystyle{tfnlm}
\bibliography{GBari_a4wd}

\newpage

\appendix

\section{List of Acronyms}
\label{sec:appendix_acronyms}
\printglossaries

\section{Training parameters}    % Each appendix must have a short title

\begin{table}[!h]
\begin{center}
\begin{tabular}{cc}
Parameter & Value \\\hline
Reference Speed ($v^{ref}$) & $20$ \\
Action scaling (\texttt{$p^{sc}$}) & $15$\\
Bound for scaled action (\texttt{$p^{bnd}$}) & $1.2$\\
\multicolumn{1}{l}{Termination reward components ($r^{ter}_{t}$):} & \\
Reaching the finish (\texttt{dist\_episode\ >\ 3900}) & $+100$\\
Leaving the track (\texttt{|track\_pos|\ >\ 1.2}) & $-10$\\
Turning back (\texttt{angle\ <\ 0}) & $-10$\\
Damage the car (\texttt{damage\ >\ 0}) & $-10$\\
Progress backwards (\texttt{speed\ <\ 0}) & $-10$\\
Low prog. (\texttt{timestep\ >\ 500\ AND\ ep\_reward < 0}) & $-10$\\ \hline
\end{tabular}
\caption{Parameters for the reward function in (eq\ref{eq2})}
\label{tab:reward}
\end{center}
\end{table}

\begin{table}[!h]
\begin{center}
\begin{tabular}{cc}
Parameter & Value \\\hline
Learning rate - with decay  &   $[1:2.5, 0: 0.5]\cdot10^{-4}$ \\
Maximum training steps & $1.5\cdot10^9$ \\
Discount factor  &     $0.995$ \\
Entropy coefficient  &   $0.01$ \\
Value function coefficient  &   $0.5$ \\
Policy clip range  &   $0.2$ \\
Value function clip range  &   $0.2$ \\
Environment instances & $24$ \\
Batch size & $512$ \\
Agent-to-Environment timestep ($ts$) & $0.05$ sec\\
% \multicolumn{1}{l}{TORCS parameters: } & \\
Used track name  &   brondehach \\
Used car model name  &     Car7 \\
\texttt{ASR\_ON} & False \\
\texttt{ASR\_ON} & False \\
Timestep in physics simulation & $0.002$ sec\\ \hline
\end{tabular}
\caption{Training parameters for Stable Baselines PPO, and the TORCS simulator environment}
\label{tab:hyper}
\end{center}
\end{table}

\begin{table}[!h]
\begin{center}
\begin{tabular}{ccc}
Measure & Description & Scale\\\hline
\texttt{LiDaR} & 17 ray, in-plane lidar signal [m] & /300 \\
\texttt{Wheel Speeds} & 4 wheel speed signals [m/s] & /80 \\
% \texttt{DIST\_EPISODE}  & \pbox{5cm}{Travelled distance along the centerline} \\
\texttt{Episode dist.}  & Distance along the centerline [m] & \\
% \texttt{CUR\_LAP\_TIME}  &  Current lap time \\
\texttt{Angle}  &  Car x axis angle to track centerline [rad] & /$\pi$\\
% \texttt{TRACK\_POS}  &  Lateral Position on the track \\
\texttt{Speed x} & CG speed in longitudinal direction [m/s] & /300\\
\texttt{Speed y} & CG speed in lateral direction [m/s] & /300 \\
% \texttt{PITCH} & Pitch Angle \\
% \texttt{PITCH\_RATE} & Pitch Rate \\
% \texttt{ROLL} & Roll Angle \\
% \texttt{ROLL\_RATE} & Roll Rate \\
% \texttt{YAW} & Yaw Angle \\
\texttt{Yawrate} & Yaw Rate [rad/s]  & /$\pi$ \\
\texttt{x Accel.} & Longitudinal Acceleration [m/ss] & /100\\
\texttt{y Accel.} & Lateral Acceleration [m/ss] & /100\\
% \texttt{SPIN\_VEL\_FL} & Front Left Wheel spin velocity \\
% \texttt{SPIN\_VEL\_FR} & Front Right Wheel spin velocity \\
% \texttt{SPIN\_VEL\_RL} & Rear Left Wheel spin velocity\\
% \texttt{SPIN\_VEL\_RR} & Rear Right Wheel spin velocity \\ 
\hline
\end{tabular}
\caption{Observation vector elements}
\label{tab:observ}
\end{center}
\end{table}

\newpage
\section{Turn-by-turn Figures}

\begin{figure}[!h]
\centering
\resizebox*{14cm}{!}{\includegraphics{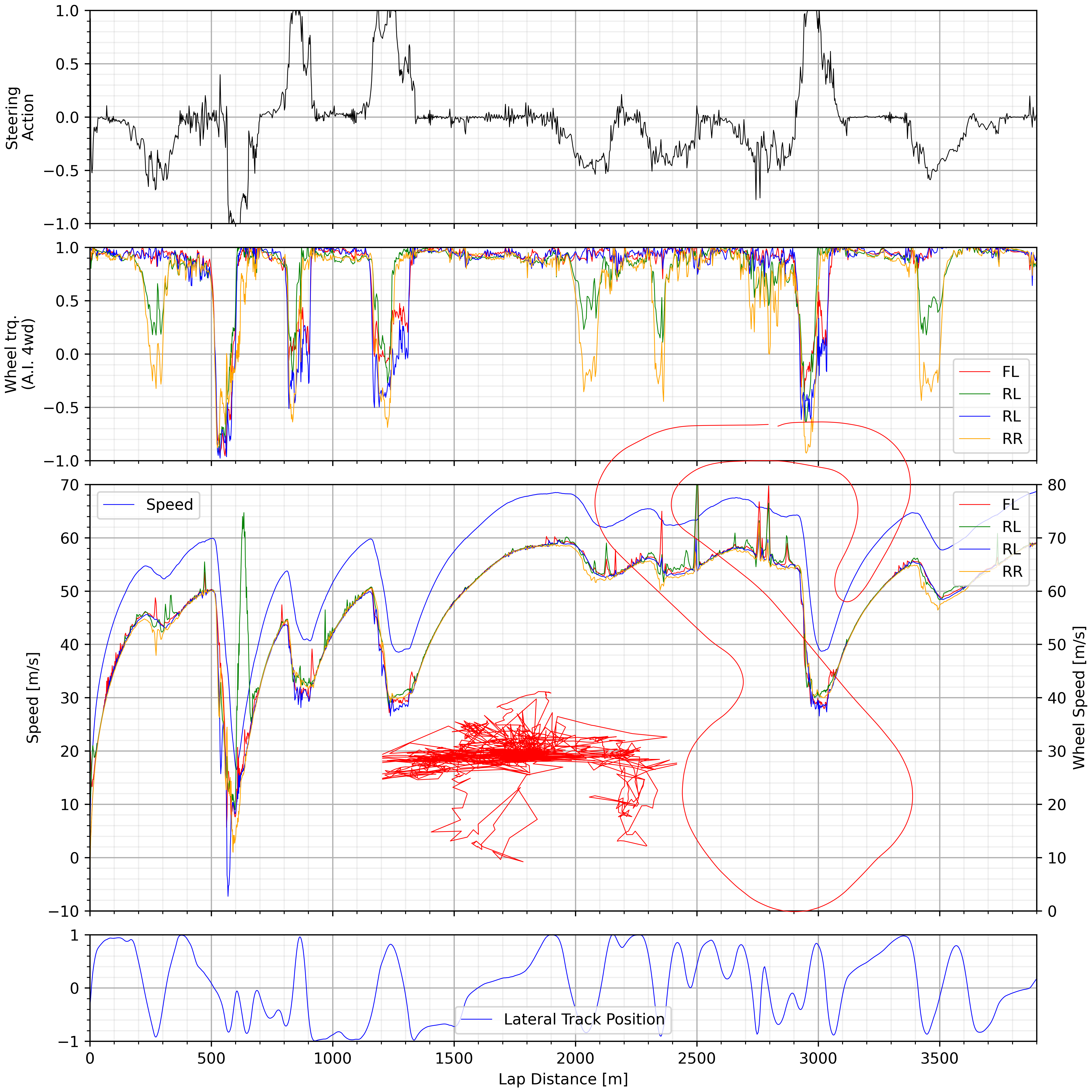}}
\caption{Plot showing a full lap driven by the trained agent controlling wheel torques and steering} 
 \label{fig:fulllap}
\end{figure}

\begin{figure}[!h]
\centering
% \resizebox*{12cm}{10.2cm}{\includegraphics{a4wdzoomT4m.png}}
\resizebox*{14cm}{!}{\includegraphics{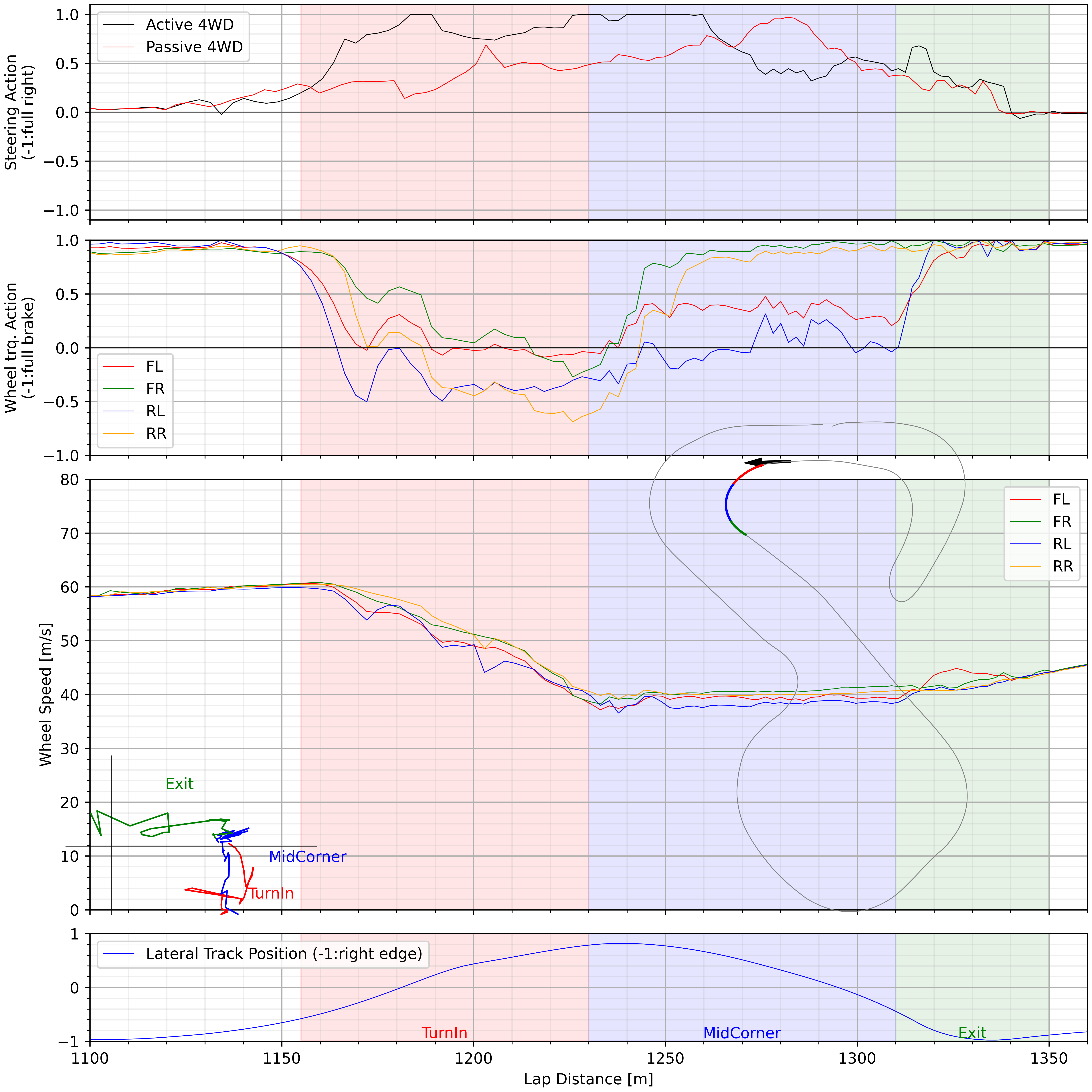}}
\caption{Diagrams for showing driving behavior in Turn4} 
 \label{fig:4wdzoomT4}
\end{figure}

\begin{figure}[!h]
\centering
% \resizebox*{12cm}{10.2cm}{\includegraphics{a4wdzoom567comp.png}}
\resizebox*{14cm}{!}{\includegraphics{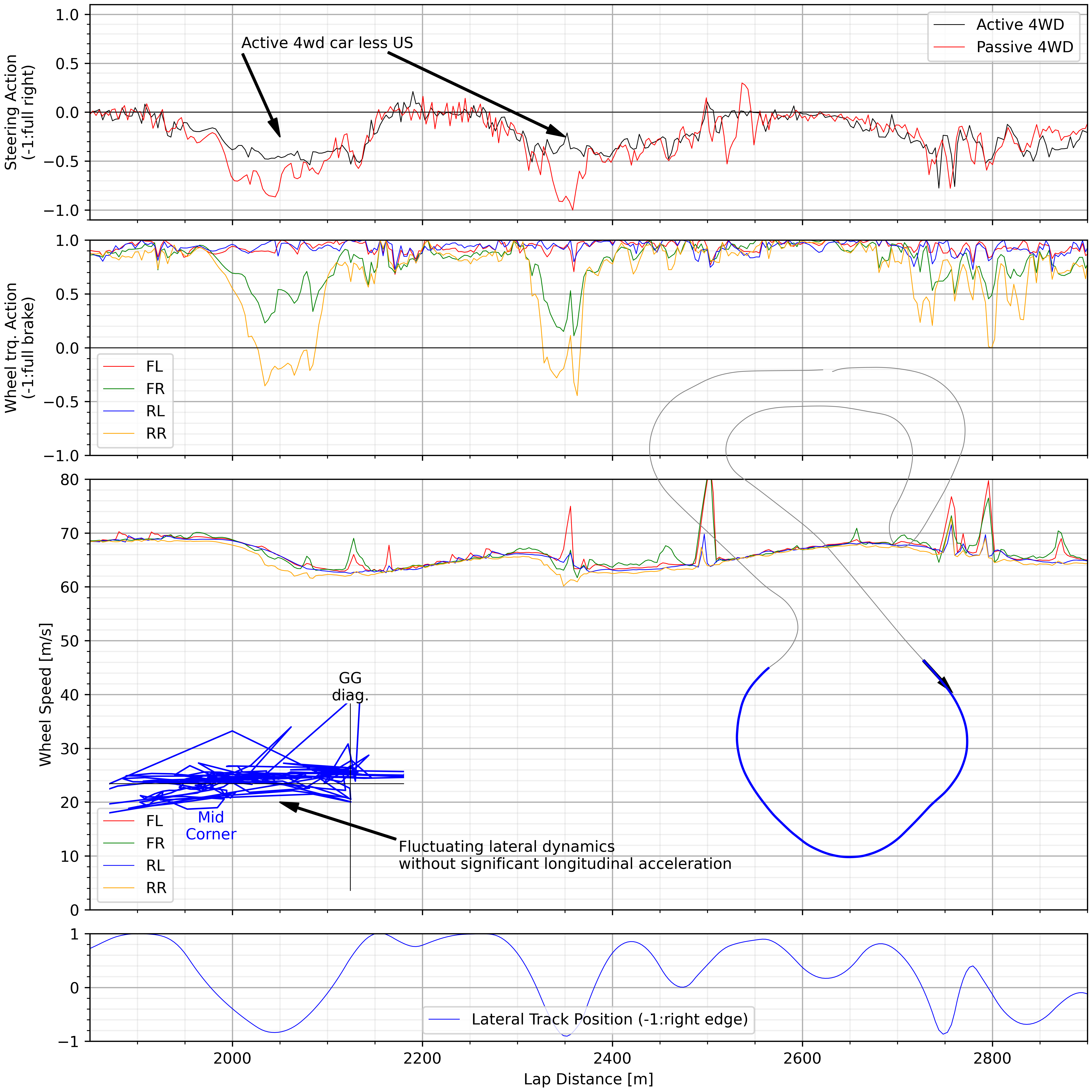}}
\caption{Diagrams for showing driving behavior in Turn5, 6, and 7. These turns represent a so called corner sequence, where corners are closely linked, the way to navigate one corner directly affects how to approach and exit the next.} 
 \label{fig:4wdzoomT567}
\end{figure}

\begin{figure}[!h]
\centering
\resizebox*{14cm}{!}{\includegraphics{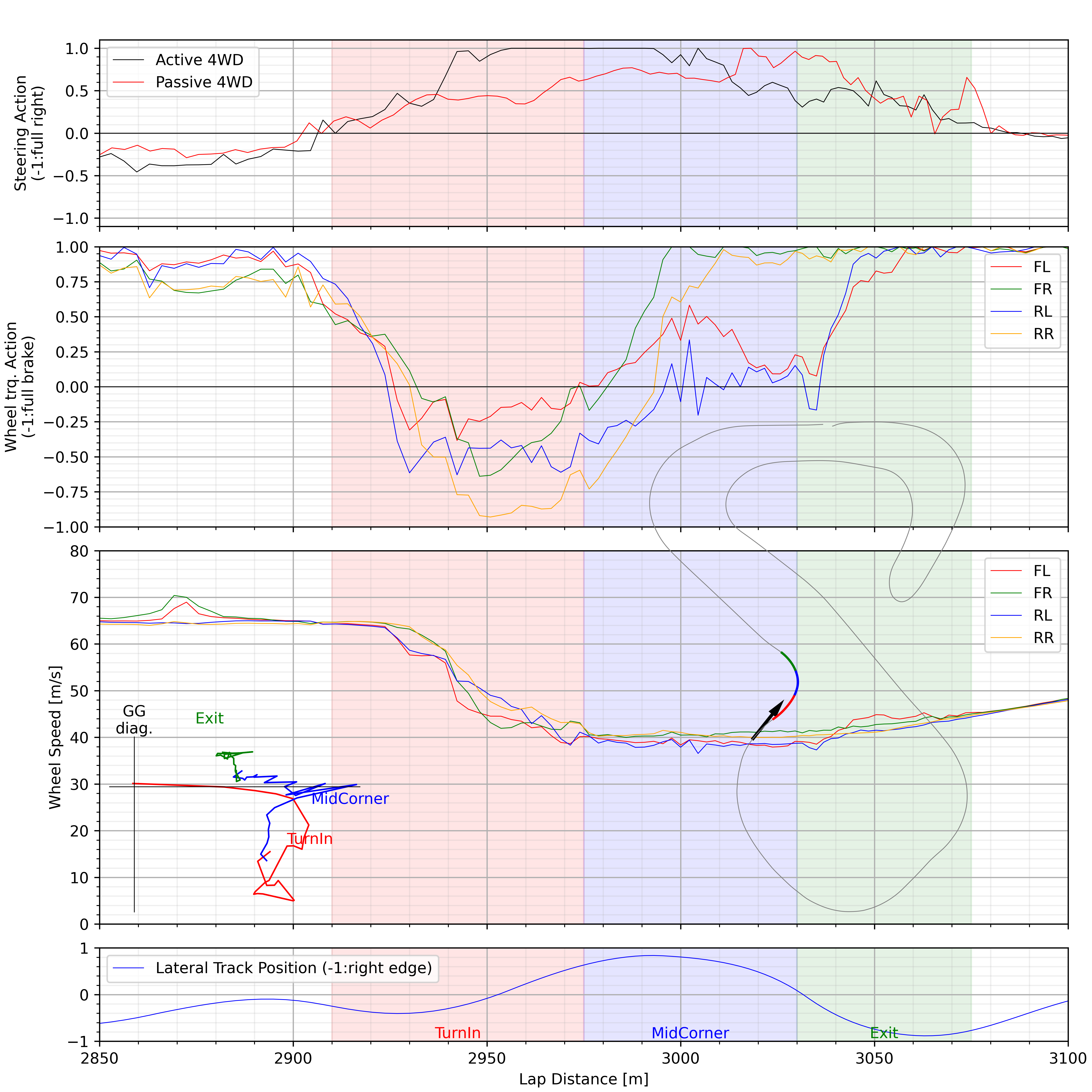}}
\caption{Diagrams for showing driving behavior in 2nd to last corner} 
 \label{fig:4wd2nd2last}
\end{figure}

\begin{figure}[!h]
\centering
\resizebox*{14cm}{!}{\includegraphics{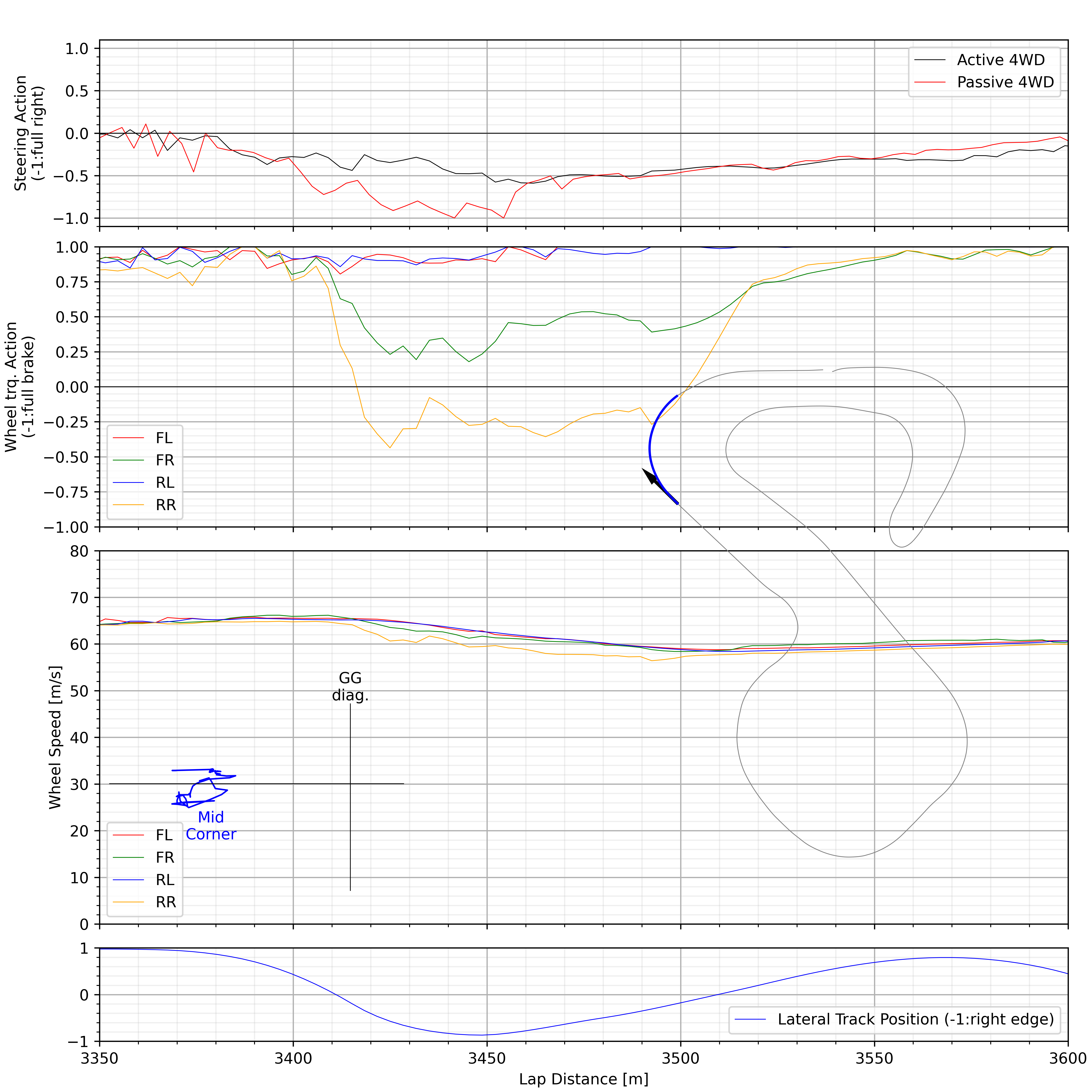}}
\caption{Diagrams for showing driving behavior in the last corner} 
 \label{fig:4wdlast}
\end{figure}

\end{document}